\documentclass[review]{elsarticle}

\usepackage{lineno,hyperref}
\modulolinenumbers[5]
\usepackage{times}
\usepackage{epsfig}
\usepackage{graphicx}
\usepackage{amsmath}
\usepackage{amssymb}
\usepackage{subcaption}
\usepackage{diagbox}
\usepackage{array}
\usepackage{adjustbox}

\journal{SI on Biometrics in the wild, Image and Vision Computing}

\newcommand{\name}{UR2D}
\newcommand{\lmdd}{X_{2D}}
\newcommand{\lmddd}{X_{3D}}
\newcommand{\proj}{P}

\newcommand{\etal}{\textit{et al.}}
\newcommand{\etc}{etc}
\newcommand{\eg}{\textit{e.g.}}
\begin{document}

\begin{frontmatter}

\title{\mbox{When 3D-Aided 2D Face Recognition Meets Deep Learning:} \\An extended UR2D for Pose-Invariant Face Recognition}

%\author{Xiang Xu, Pengfei Dou, Ha A. Le, Yuhang Wu, Miloud Aqqa, Ioannis A. Kakadiaris\corref{mycorrespondingauthor}}
\author{Xiang Xu}
%\ead{xxu18@central.uh.edu}

\author{Pengfei Dou}
%\ead{pdou@central.uh.edu}

\author{Ha A. Le}
%\ead{hale4@central.uh.edu}

\author{Ioannis A. Kakadiaris\corref{mycorrespondingauthor}}
\ead{ikakadia@central.uh.edu}
\cortext[mycorrespondingauthor]{Corresponding author}
\address{Computational Biomedicine Lab\\University of Houston\\4800 Calhoun Rd. Houston, TX, USA}

%\ead{\{xxu18, pdou, hale4, ywu35, maqqa, ikakadia\}@central.uh.edu}
%\ead{ikakadia@central.uh.edu}


\begin{abstract}
% A few well-developed face recognition pipelines have been reported in recent years.
Most of the face recognition works focus on specific modules or demonstrate a research idea. 
This paper presents a pose-invariant 3D-aided 2D face recognition system (\name) that is robust to pose variations as large as 90$^{\circ}$ by leveraging deep learning technology.
The architecture and the interface of \name \ are described, and each module is introduced in detail. 
Extensive experiments are conducted on the UHDB31 and IJB-A, demonstrating that \name \ outperforms existing 2D face recognition systems such as VGG-Face, FaceNet, and a commercial off-the-shelf software (COTS) by at least $9\%$ on the UHDB31 dataset and $3\%$ on the IJB-A dataset on average in face identification tasks.
%This paper extends \name \ by generating a template from several images for a subject and optimize the GPU usage in this paper from our conference paper \cite{Xu_2017_17643}. 
\name \ also achieves state-of-the-art performance of $85\%$ on the IJB-A dataset by comparing the Rank-1 accuracy score from template matching.
It fills a gap by providing a 3D-aided 2D face recognition system that has compatible results with 2D face recognition systems using deep learning techniques.
\end{abstract}


\end{frontmatter}

%\linenumbers

\section{Introduction}

\begin{figure}[tb]
\begin{subfigure}{\linewidth}
\includegraphics[width=\linewidth]{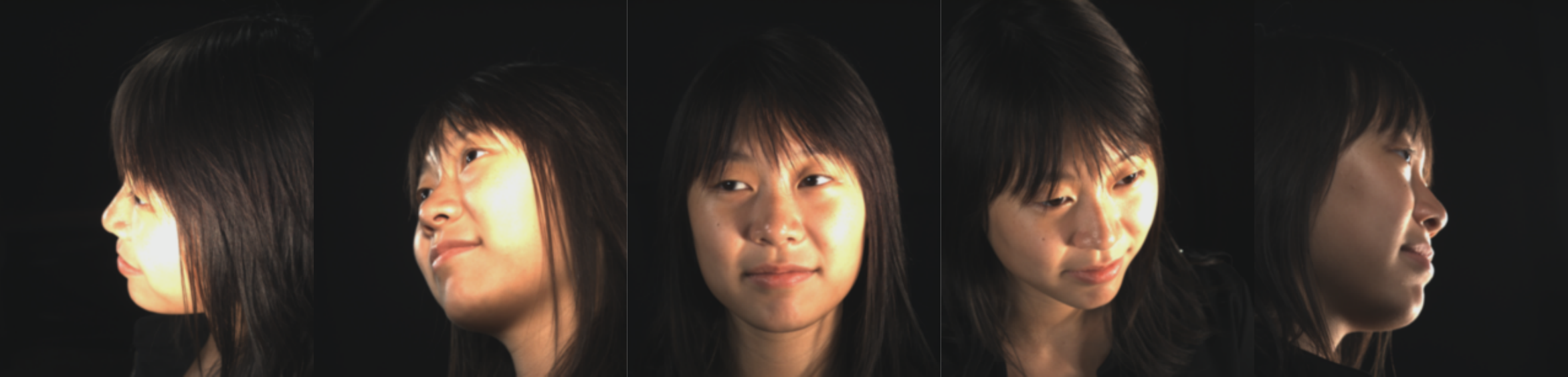}
%\caption{}
\end{subfigure}
~
\begin{subfigure}{\linewidth}
\includegraphics[width=\linewidth]{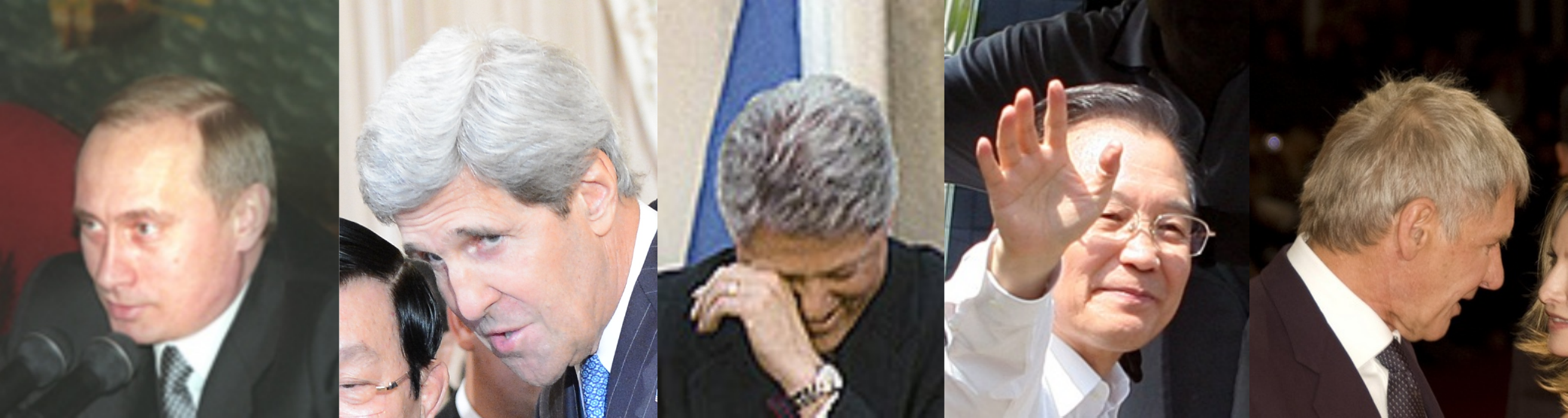}
%\caption{}
\end{subfigure}
\caption{Depiction of existing pose problem from selected samples. Distribution of yaw angles are from $-90^{\circ}$ to $+90^{\circ}$ in (T) constrained dataset UHDB31 \cite{Le_2017_17704} and in (B) the wild dataset IJB-A \cite{Klare_2015_17419}.}
\label{fig:pose}
\end{figure}

Face recognition is an application in which the computer either classifies human identity according to the face (face identification) or verifies whether two images belong to the same subject (face verification). 
A common face recognition system has two steps: enrollment and matching. 
Specifically, in the enrollment stage, features are obtained from a facial image or a set of images to obtain a signature or a template for each subject. 
The enrollment usually has three steps: (i) face detection, (ii) face alignment, and (iii) signature generation.
In the matching stage, these signatures are compared to obtain a distance for the identification or verification problem.
Recently, face recognition technology has significantly advanced by the deployment of deep learning technology, especially using Convolutional Neural Networks (CNN). 
Pure 2D face recognition (2D-FR) systems have achieved human performance or even better. 
DeepFace, proposed by Taigman \etal \ \cite{Taigman_2014_15190}, first reported performance on the Labeled Faces in the Wild (LFW) standard benchmark \cite{Huang_2008_13354} that was better than human efforts.
FaceNet, proposed by Schroff \etal \  \cite{Schroff_2015_16462}, used triplet loss to train a deep neural network using 200 million labeled faces, and obtained a performance of $99.63\%$ verification accuracy on the same dataset.
The success of deep learning techniques in face recognition indeed relies on the following four aspects: (i) a large amount of data either from public datasets such as WebFace \cite{Yi_2014_17446} and Ms-Celeb-1M \cite{Guo_2016_17416}, or private datasets, (ii) advanced network architecture such as VGG \cite{Parkhi_2015_16638} and ResNet \cite{He_2016_17189}, (iii) discriminative learning approaches such as Triplet Loss \cite{Schroff_2015_16462}, Center Loss \cite{Wen_2016_17264}, Range Loss \cite{Zhang_2016_17759}, SphereFace \cite{Liu_2017_17761}, and (iv) regularization methods such as Noisy Softmax \cite{Chen_2017_17762}.

However, face recognition is still not a solved problem in real-world conditions. 
Some datasets, such as LFW, use Viola-Jones face detector, which is not designed to work in the whole pose distribution from $-90^{\circ}$ to $+90^{\circ}$.
In an unconstrained scenario, especially using surveillance camera, there is a plethora of images with large variations in head pose, expression, illumination, and occlusions. 
To overcome these challenges, a 3D face model can be applied to assist a 2D face recognition. A 3D facial model is intrinsically invariant to pose and illumination.
To use a 3D face model, a model should is fitted on the facial images and a 3D-2D projection matrix is estimated. 
With the help of a projection matrix and fitted 3D model, it is easy to rotate the face out-plane and align the input images from any arbitrary large pose positions to the frontal position for the feature extraction and signature matching.

In the last few years, researchers focused on the 2D face recognition from pure 2D image view and have developed numerous loss function approaches to learn the discriminative features from the different poses. A \textbf{limited} number of 3D-aided 2D face recognition systems (3D2D-FR) have been developed using the 3D model to help align 2D images.
Kakadiaris \etal \ \cite{Kakadiaris_2017_13356} proposed a pose and illumination invariant system which frontalized the face image using annotated face model (AFM). 
% However, the performance of these systems is far from being compatible with 2D-FR using deep learning.
Hu \etal \ \cite{Hu_2016_17415} proposed a unified 3D morphable model (U-3DMM) which has additional PCA subspace for perturbation.

To address the problem mentioned above, this paper presents a 3D-aided 2D face recognition system called \name \, which significantly improves face recognition performance using the AFM and deep learning technology, especially in large pose scenarios.
There is enormous demand \cite{Ding_2016_16547} for pose-invariant face recognition systems because frontal face recognition can be considered as a solved problem.
%Another main purpose is to accelerate face-related research in the group by providing ample tools and applications.

\name \ consists of several independent modules: face detection, landmark detection, 3D model reconstruction, pose estimation, lifting texture, signature generation, and signature/template matching. Despite face detection methods, all other methods are developed in the Computational Biomedicine Lab.
It provides sufficient tools and interfaces to use different sub-modules designed in the system.
The core code is written in efficient C++, which provides bindings to Python.
The system leverages several open-sourced libraries such as 
OpenCV \cite{noauthor__17451},
glog \cite{noauthor__17452}, 
gflags \cite{noauthor__17453}, 
pugixml \cite{noauthor__17454}, 
JSON for modern C++ \cite{noauthor__17455},
and Caffe \cite{Jia_2014_16330}.

In \name, after detecting the face and 2D landmarks from image, a 3D model is constructed from a 2D image or several 2D images. By estimating the 3D-2D projection matrix, the correspondence between the 3D model and 2D image can be computed. 
Then, a 3D model is used to help frontalize the face. 
The pose-robust features and occlusion encodings are extracted to represent the face. 
For matching, we use cosine similarity to compute the similarity between two signature vectors.

In summary, this paper extends Xu \etal \ \cite{Xu_2017_17643} and make the following contributions:
\begin{itemize}
\item A brief survey of recent face recognition pipeline and each module are summarized;
\item A pose-invariant 3D-aided 2D face recognition system using deep learning is developed. The intrinsic value of a 3D model is explored to frontalize the face, and the pose-invariant features are extracted for representation. We demonstrate results that a 3D-aided 2D face recognition system exhibits a performance that is comparable to a 2D only FR system. Our face recognition results outperform the VGG-Face, FaceNet, and COTS by at least $9\%$ on the UHDB31 dataset and $3\%$ on the IJB-A dataset on average.
In addition, we demonstrate that \name \ can generate template signatures from multiple images and achieve state-of-the-art performance of $85\%$ on the IJB-A dataset.
\end{itemize}

%We fill the gap that there is no modern 3D2D-FR system exists that can be directly used in real life in the recent years.
%\end{itemize}

The rest of the paper is organized as follows: modern face recognition systems are reviewed in Sec.~\ref{sec:related}. In Sec.~\ref{sec:intro}, the architecture of \name \ and its functionalities are discussed. In Sec.~\ref{sec:fr}, each module separately is introduced in detail. Detailed evaluations on the indoor and in-the-wild datasets are reported in Sec.~\ref{sec:exp}. 
%In addition, we discuss some future works in Sec.~\ref{sec:future}.

\section{Related work}
\label{sec:related}
We divide the current existing face-related work into two categories:
In Sec.~\ref{sec:rw_module}, we discuss detailed recent related work for each module in the common face recognition pipeline from an academic view.
System level papers about the implementation are discussed in Sec.~\ref{sec:rw_system} .

\subsection{Modules}
\label{sec:rw_module}
\textbf{Face Detection}:
Face detection is the first step, as well as the most studied topic, in the face recognition domain. 
Zefeiriou \etal \ \cite{Zafeiriou_2015_17448} presented a comprehensive survey on this topic. They divided the approaches into two categories: rigid template-based methods, and deformable-parts-models-based methods.
In addition to the methods summarized in \cite{Zafeiriou_2015_17448}, the approaches of object detection under the regions with a convolutional neural network (R-CNN) framework \cite{Girshick_2014_17449} have been well developed.
Some techniques can be directly integrated to face detection \cite{Jiang_2017_17450}. 
Li \etal \ \cite{Li_2016_17296} used a 3D mean face model and divided the face into ten parts. They joined face proposals into a single R-CNN model.
The approach proposed by Hu and Ramanan \cite{Hu_2017_17443} explored context and resolution of images to fine-tune the residual networks (ResNet) \cite{He_2016_17161}, which was demonstrated to detect a face as small as three pixels.
Despite the two-stage face detectors above using proposal and classification technique, single-stage detectors have also been developed.
SSD \cite{Liu_2016_17507} and YOLO \cite{Redmon_2017_17763} classify a fixed grid of boxes and learn regression functions to map to the objects simultaneously. 
Lin \etal \ \cite{Lin_2017_17764} address the issue that the performance of single-stage detectors are not as strong as two-stage detectors because of unbalanced positive and negative samples. With focal loss, they also trained state-of-the-art single-stage object detector.
Very recently, SSH has been proposed by Najibi \etal \ \cite{Najibi_2017_17765} using multi-task loss for both classification and regression in the network.

\textbf{Face Alignment}:
Face alignment refers to aligning the face image to a specific position.
Usually, researchers include landmark detection in this topic.
\mbox{Jin and Tan \cite{Jin_2017_17152}} summarized the categories of popular approaches for this task.
Cascaded regression was a major trend in this topic and classification frameworks tend to be popular recently.
Zhu \etal \ \cite{Zhu_2015_16323} searched for similar shapes from exemplars and regressed the shapes by using SIFT features and updating the probability of shapes.
An ensemble of random ferns \cite{Xu_2016_16539} are used to learn the local binary discriminative features.
Xu and Kakadiaris \cite{Xu_2017_17394} proposed to jointly learn head pose estimation and face alignment tasks in a single framework (JFA) using global and local CNN features.
Some researchers treat the face alignment task as a classification problem.
KEPLER \cite{Kumar_2017_17402} joined CNN features from different layers and captured the response map to localize the landmarks.
Wu \etal \ \cite{Wu_2017_17445} proposed the GoDP algorithm to localize landmarks under a fully convolutional network (FCN) framework by exploring two-pathway information.
Some recent works use generative adversarial networks (GAN) to frontalize the face \cite{Huang_2017_17458, Yin_2017_17457, Tran_2017_17705}.
Huang \etal \ \cite{Huang_2017_17458} used two-pathway GAN (TP-GAN) for photo-realistic frontal synthesis images, but kept identity and details of texture.
Yin \etal \ \cite{Yin_2017_17457} incorporated a 3D model with GAN to frontalize faces for large poses in the wild.
DR-GAN was proposed by Tran \etal \ \cite{Tran_2017_17705} to generate the frontalized face from face images under different poses. They also demonstrated the usage of GAN in face recognition.

\textbf{Signature Generation}:
An emerging topic in face recognition research is generating a discriminative representation for a subject.
When training with millions of face images using deep learning technology, many feature descriptors have been proposed recently.
Parkhi \etal \ \cite{Parkhi_2015_16638} proposed the VGG-Face descriptor within VGG-Very-Deep architectures.
Triplet loss was proposed by Schroff \etal \ \cite{Schroff_2015_16462} to train a deep neural network using 200 million labeled faces from Google.
Masi \etal \ \cite{Masi_2016_17266} developed face recognition for unconstrained environments by fine-tuning the ResNet and VGG-Face on 500K 3D rendering images.
In addition to frontalizing the face, they also rendered face images to half-profile $40^{\circ}$, and full-profile ($75^\circ$).
Masi \etal \ \cite{Masi_2016_17399} addressed the question of whether we need to collect millions of faces for training a face recognition system.
They argued that we can use synthesized images instead of real images to train the model and still obtain comparable results.
Despite triplet loss, many other loss functions have been proposed recently.
Center loss was added by Wen \etal \ \cite{Wen_2016_17264} alongside cross entropy loss to obtain more discriminative features for deep face recognition.
Range loss \cite{Zhang_2016_17759} was designed by Zhang \etal \ to train deep neural networks with a long tail distribution.
A-Softmax Loss \cite{Liu_2017_17761} was used in SphereFace and demonstrated efficiency in learning the discriminative features.
Marginal Loss \cite{Deng_2017_17766} was proposed to enhance the discriminative ability by maximizing the inter-class distances from large scale training data.

\begin{table}[t]
\begin{center}
\begin{adjustbox}{max width=\linewidth}
\begin{tabular}{|l|c|c|c|c|c|c|c|c|}
\hline
Name & Category    & Core    & Detection     & Alignment & Representation  & Matching & Modern & Active \\ \hline \hline
OpenBR \cite{Klontz_2013_14723}    &  2D     & C++ & \checkmark    & \checkmark &    \checkmark  & \checkmark    &       &       \\ \hline
FaceID1-3 \cite{Sun_2015_16639}       &  2D  &  -  &        &   &   \checkmark   & \checkmark &  \checkmark &  \\ \hline
DeepFace \cite{Taigman_2014_15190} &  2D  &  -  &     &   &    \checkmark  & \checkmark    & \checkmark      &    \\ \hline
FaceNet \cite{Schroff_2015_16462}  &  2D  &  -  &       &       &     \checkmark     & \checkmark    & \checkmark      &      \\ \hline
VGG-Face \cite{Parkhi_2015_16638}       &  2D     &  -  &         &     & \checkmark &\checkmark &   \checkmark    &       \\ \hline
OpenFace \cite{Amos_2016_16583}       &  2D     &  Torch  &    \checkmark     & \checkmark      & \checkmark &\checkmark &   \checkmark    &       \\ \hline
U-3DMM \cite{Hu_2016_17415}        & 3D2D &  -  &        &   &    \checkmark & \checkmark    & \checkmark     &    \\ \hline \hline
\name                              & 3D2D & C++ &    \checkmark     & \checkmark & \checkmark  & \checkmark    &  \checkmark & \checkmark \\ \hline
\end{tabular}
\end{adjustbox}
\end{center}
\caption{Comparison of recent existing 2D face recognition pipelines. We employ the same definition of \textit{modern} and \textit{active} made by Klontz \etal \ \cite{Klontz_2013_14723} (``-" means that this information is not provided in the paper).}
\label{tb:summary}
\end{table}

\subsection{System}
\label{sec:rw_system}
OpenCV and OpenBR are some well known open-source computer vision and pattern recognition libraries.
However, the eigenface algorithm in OpenCV is out-of-date. 
OpenBR has not been updated since 9/29/2015.
Both libraries only support nearly frontal face recognition, since the face detector can only detect the frontal face.
OpenFace is an open-source implementation of FaceNet \cite{Schroff_2015_16462} by Amos \etal \ \cite{Amos_2016_16583} using Python and Torch, which provides four demos for usage.
OpenFace applied Dlib face detector and landmark detector to do the pre-processing, which is better than OpenBR.
There is another official Tensorflow implementation of FaceNet in which the authors use MTCNN \cite{Zhang_2016_17265} to detect and align face, which boosts performance speed and detection accuracy.

To the best of our knowledge, there is a limited amount of well-designed system papers.
Most face-related papers focus on different sub-modules or the research of face representations.
The comparison of recent existing 2D face recognition systems is presented in Tab.~\ref{tb:summary} including the research on face representation.

\section{System Design}
\label{sec:intro}
\name \ is a 3D-aided 2D face recognition system designed for pose-invariant face recognition. 
Moreover, this system is suitable for face-related research, and can fast pre-process images, provide baselines, plot the results, and support further development.

\subsection{System requirements}
\name \ is written in clean and efficient C++, which is developed on a Linux platform (Ubuntu system).
It requires GCC 4.9 or above for compilation. 
It leverages a list of open-sourced libraries and tools such as CMake, Boost, OpenCV, gflags, glog, puxixml, JSON, and Caffe.
Most of the dependencies are available in the Ubuntu repository except Caffe. 
Therefore, to install dependencies, it only requires installing Caffe manually.

\begin{figure}[thb]
\begin{center}
\includegraphics[width=\linewidth]{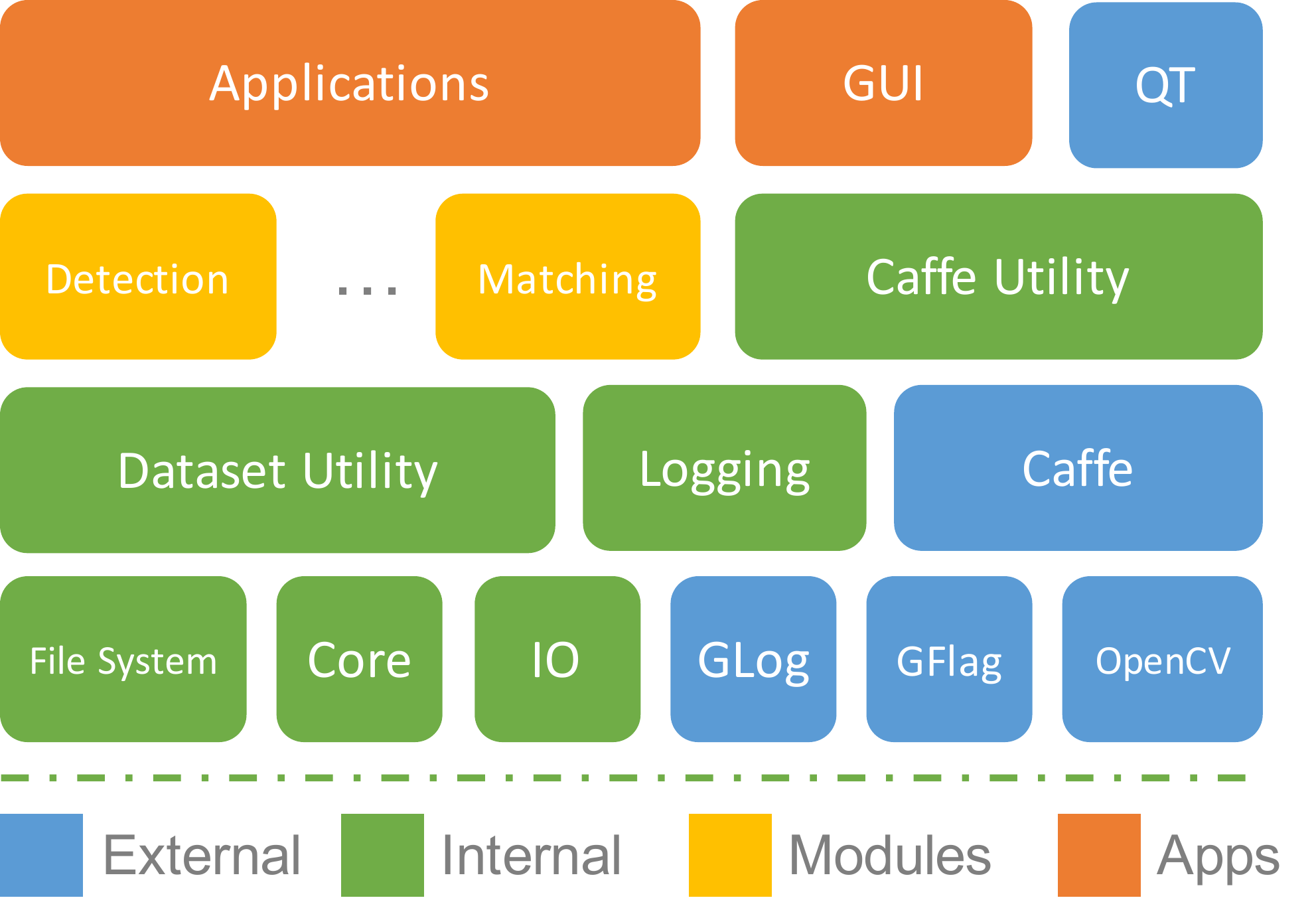}
\end{center}
\caption{Depiction of \name's architecture. In addition to external libraries, it includes some other base libraries to process files, use CUDA, manage the data files, \etc. Based on these basic libraries, high level APIs were implemented by calling function from each module. Based on \name's SDK, it is easier to write various applications for different purposes. Also, we created the GUIs to demonstrate our \name.}
\label{fig:arch}
\end{figure}

\subsection{Architecture Overview}
Figure~\ref{fig:arch} illustrates the architecture of \name, which explicitly illustrates modules and functionality. 
The blue blocks are external shared libraries.
The other three components belong to our system.
As a base of the software, green blocks provide the basic functions.
The algorithm modules are constructed as high-level APIs.
The applications and GUIs are the top of the software and are built by combining these APIs.
The users can directly call these applications and obtain the results.
The advantages of this architecture are that it is simple and well-structured.
With full development of libraries, the system can use CPUs/GPU and other features easily.

\subsection{Data Structures}
In \name, the basic element is \texttt{File} on the disk. 
All operations or algorithms are based on the files. 
The basic data structure is \texttt{Data}, which is a hash table with pairs of keys and values. 
Both keys and values are in string type.
Unlike OpenBR \cite{Klontz_2013_14723}, to avoid saving giant data in the memory, we only keep the file path in the memory.

\subsection{Configuration}
We have two approaches to run \name. 
The first one is defining the configuration file (JSON format), which points out the datasets, input files, output directories, involving modules and their model locations, and evaluation. 
Attribute \texttt{dataset} contains the information of input dataset including the name and path. Attribute \texttt{input} contains the list of galleries and probes. Attribute \texttt{output} defines the output directories. Attribute \texttt{pipelines} defines the modules used in the pipeline.
The \texttt{pip} command line application only accepts the argument of the configuration file, which will parse the configuration file, load the models, and run defined modules.
The advantages of this approach are simplicity and flexibility.
Unlike the OpenBR framework, it does not require a detailed understanding of the option or input long arguments in the command line.
The users only need to change some values in the attributes \texttt{dataset} and \texttt{input} (\eg, set dataset directory and file to enroll), and program \texttt{pip} will generate the output they defined in this configuration file.

%\begin{figure}[tb]
%\centering
%\includegraphics[width=\linewidth]{imgs/config_sample.eps}
%\caption{Depiction of our configuration file. It is a long JSON file. We omit some parts of the file due to the space reason in this paper. Attribute \texttt{dataset} contains the information of input dataset including the name, path. Attribute \texttt{input} contains the list of galleries and probes. Attribute \texttt{output} defines the output directories. Attribute \texttt{pipelines} defines the modules used in the pipeline. In this case, \texttt{face detection} module will use headhunter face detector and no evaluation function will be processed.}
%\label{fig:config}
%\end{figure}
%
%\begin{figure}[tb]
%\centering
%\includegraphics[width=\linewidth]{imgs/enroll.png}
%\caption{Enrolling a list of files and generates signature from IJB-A gallery split 1.}
%\label{fig:enroll}
%\end{figure}

\subsection{Command Line Interface}
To make full use of SDK of \name, we created some corresponding applications to run each module.
All applications accept the file list (text or csv file by default, which includes tag at the top line), a folder, or a single image.
The \texttt{IO} system will load the data in the memory and process the data according to the data list.

The arguments specify the location of the input file/directory and where the output should be saved.
\name's enrollment is executed and generates signatures to the output directory.
The path of the signature is recorded in the \texttt{Data}.
By calling the API from \texttt{IO} system, the list of \texttt{Data} will be written to the file (default is in \texttt{.csv} format).

%\subsection{Build options}
%When building the project by Cmake, several options are provided such as \texttt{USE\_CUDA, USE\_CAFFE, USE\_OPENMP, BUILD\_DOC, BUILD\_PYTHON}, \etc.
%By default, the program set \texttt{USE\_CUDA} and \texttt{BUILD\_GUI} to \texttt{OFF} and the others are \texttt{ON} for generalization of different system configurations.
%This option name explicitly describes the meaning of the option.
%For example, 
%when \texttt{USE\_CUDA} is turned on, the library will be in the GPU mode, which will accelerate the program.
%When \texttt{BUILD\_GUI} is turned on, the library will find the QT5 library and build the demo GUI for usage.

\section{Face Recognition}
\label{sec:fr}

\begin{figure*}[htb]
\begin{center}
\includegraphics[width=\linewidth]{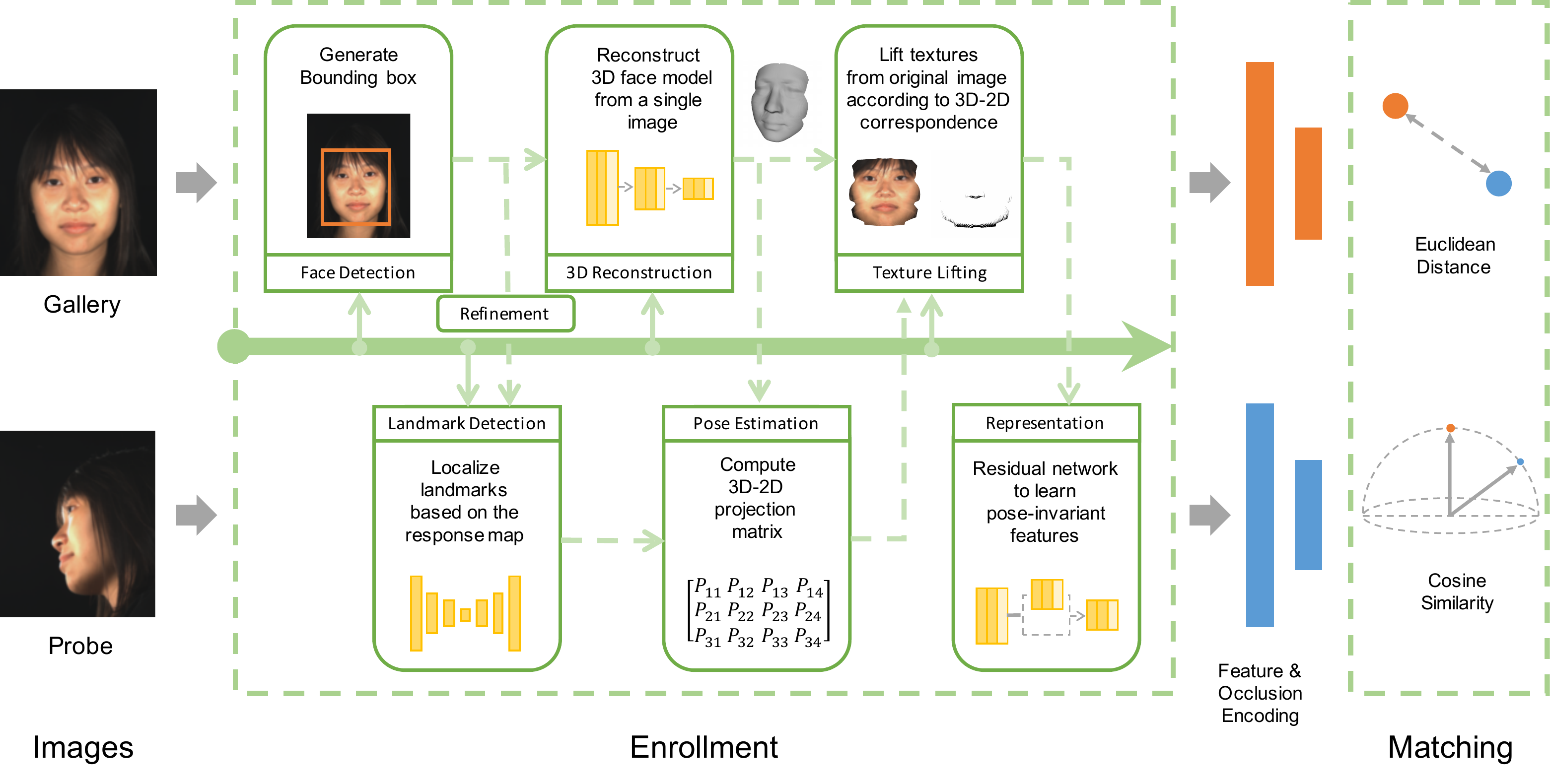}
\end{center}
\caption{Depiction of the whole pipeline (follow the arrow in the middle) of \name. The rounded \mbox{rectangles} represent different modules. Dashed arrows represent the workflow. The enrollment encompasses the modules listed. A face is first detected and then transferred to localize landmarks. A 3D model is constructed directly from a 2D image with a bounding box. With 2D landmarks and a 3D model, a 3D-2D projection matrix can be estimated. The frontalized image and occlusion map are generated according to the 3D model and projection matrix. The pose robust features are extracted from these images along with occlusion encoding. The matching step computes features from visible parts and outputs a similarity score.}
\label{fig:pip}
\end{figure*}

Figure~\ref{fig:pip} depicts the overview of enrollment in the \name, which contains face detection, face alignment, 3D face reconstruction, pose estimation, texture lifting, and signature generation.

\subsection{Face Detection}
A serious problem in OpenBR \cite{Klontz_2013_14723}, OpenFace \cite{Amos_2016_16583}, and even the commercial off-the-shelf face recognition software (COTS) is the face detection rate. 
OpenBR only supports OpenCV frontal face detector. 
OpenFace also supports Dlib \cite{King_2009_17112} face detector.
However, in recent years, many face detection algorithms have been developed \cite{Farfade_2015_16581, Li_2016_17296, Hu_2017_17443} with deep learning technology to support multi-view images.

To detect the face in multi-view poses, some modern detectors such as Headhunter \cite{Mathias_2014_17444} and DDFD \cite{Farfade_2015_16581}, and Dlib-DNN face detector are supported in our system. 
Mathias \etal \ \cite{Mathias_2014_17444} trained Headhunter by using multi-scale templates.
DDFD face detector is proposed by Farfade \etal \ \cite{Farfade_2015_16581} by fine-tuning AlexNet \cite{Krizhevsky_2012_17120} and using non-maximum suppression (NMS-max, NMS-avg).

To support different face detectors for downstream modules, we perform the bounding box regression on detected bounding box to reduce the variations of the bounding box.
The first advantage of this approach is that we do not need to re-train or fine-tune the models for downstream modules after switching the face detector.
The second advantage is that this approach provides a more robust bounding box for the landmark localization module.

\subsection{Landmark Localization}
To detect face landmarks, we use GoDP proposed by Wu \etal \ \cite{Wu_2017_17445}, which is demonstrated to be robust to pose variations.
GoDP landmark detector relies on confidence maps generated by a fully convolutional network. 
A confidence map is generated for each landmark to indicate the possibility of a landmark appearing at a specific location in the original image. 
The prediction is made by simply selecting the location that has the maximum response in the confidence map. 
This winner-take-all strategy helps to suppress false alarms generated by background regions and improves the robustness of the algorithm under large head pose variations. 
Compared to other confidence-map-based landmark detectors, the novel architecture of GoDP merges the information of the deep and shallow layers based on a new loss function, increases the resolution and discrimination of the confidence maps, and achieves state-of-the-art results on multiple challenging face alignment databases. 

\subsection{3D Reconstruction of Facial Shape}
To reconstruct the 3D facial shape of the input 2D image, we integrate into our pipeline the E2FAR algorithm proposed by Dou \etal \ \cite{Dou_2017_17421}. 
It uses a subspace model to represent a 3D AFM as a parameter vector and employs CNN to estimate the optimal parameter values from a single 2D image. 
To train the deep neural network, a large set of synthetic 2D and 3D data has been created using the 3D rendering of randomly generated AFMs. 
To improve the robustness to illumination variation, the deep neural network is pre-trained on real facial images and fine-tuned on the synthetic data. 
Compared with existing work, it is more efficient due to its end-to-end architecture, which requires a single feed-forward operation to predict the model parameters. 
Moreover, it only relies on face detection to localize the facial region of interest on the image. 
As a result, compared with landmark-based approaches, it is more robust to the pose variation that can degrade landmark detection accuracy.  

\subsection{Pose estimation}
Given 2D landmarks $\lmdd$ obtained from landmark detection and 3D landmarks $\lmddd$ obtained from a 3D model, the transformation matrix $\proj$ can be estimated by solving a least-squares problem as follows:
\begin{equation}
\min_{P} ||\lmdd - \proj \lmddd||_{2}^{2} + \lambda||P||_{2}^{2}.
\end{equation}

In our implementation, we use the Levenberg-Marquardt algorithm, also known as DLS, to solve this equation.
%In \name, we use Eigen\footnote{\url{http://eigen.tuxfamily.org}} to solve this approximation problem.

\subsection{Texture Lifting}
Facial texture lifting is a technique first proposed by \mbox{Kakadiaris \etal \ \cite{Kakadiaris_2017_13356}}, which lifts the pixel values from the original 2D images to a UV map.
Given the 3D-2D projection matrix $P$, 3D AFM model $M$, and original image $I$, it first generates the geometry image $G$, each pixel of which captures the information of an existing or interpolated vertex on the 3D AFM surface. 
With $G$, a set of 2D coordinates referring to the pixels on an original 2D facial image is computed.
In this way, the facial appearance is lifted and represented into a new texture image $T$. 
A 3D model $M$ and Z-Buffer technique are used to estimate the occlusion status for each pixel. This process generates an occlusion mask $Z$.

This module has the following two advantages: It generates the frontal normalized face images, which is convenient for feature extraction and comparison.
Second, it generates occlusion masks, which identify the parts of the face images that are occluded, providing the evidence to exclude the face regions.

\subsection{Signatures}
To improve the performance of face recognition in matching non-frontal facial images, we integrate into our pipeline the algorithm proposed by Dou et al. \cite{Dou_2015_16379} for extracting Pose-Robust Face Signature (PRFS), a part-based face representation with discriminative local facial features and explicit pose and self-occlusion encoding. 
The facial texture $T$ and the self-occlusion mask $Z$ are first divided into multiple local patches. 
Then, on each local patch, discriminative features are extracted and self-occlusion encoding is computed. The ensemble of local features, each enhanced by the self-occlusion encoding, forms the pose-robust face signature. We use two types of local features, namely the DFD feature proposed by Lei \etal \cite{Lei_2014_16175} and a deep feature we trained by following Wen \etal \cite{Wen_2016_17264} using center loss. To train the DFD feature, we use a small subset of the FRGC2 database that consists of 907 frontal facial images of 109 subjects. We divide the facial texture into 64 non-overlapping patches and train a DFD feature extractor for each local patch separately. To train the deep feature, the CASIA WebFace dataset \cite{Yi_2014_17446} is used as training data. We divide the facial texture into 8 partially-overlapping patches and train a deep neural network for each local patch separately. In this paper, we call the face signature with the DFD feature PRFS, and the face signature with the deep feature DPRFS.

%\subsection{Matching}
%In signature matching, local similarities are computed on non-occluded local patches and the overall similarity is the average of local similarities.
%\name \ provides the APIs to compute cosine similarity and euclidean distance and generates the score matrix.

\section{Experiments}
\label{sec:exp}

In this section, we provide a systematical and numerical analysis on two challenging datasets in both constrained and in-the-wild scenarios.
First, the datasets used to verify \name \ are introduced.
Then, a fair comparison of \name \ with VGG face descriptor (VGG-Face) and a commercial face recognition software (COTS) on these two challenging datasets is conducted for the image matching. 
In the end, the template matching experiments on IJB-A dataset was performed.

\begin{table}[thb]
\begin{center}
\begin{adjustbox}{max width=\linewidth}
\begin{tabular}{|l|c|c|c|c|c|c|}
\hline
Dataset & Images & Subjects & Environment & Poses & Illuminations & Usage\\ \hline \hline
UHDB31 & 24,255 & 77 & Constrained & 21 & 3 & 2D-2D, 3D-2D, 3D-3D face recognition\\ \hline
IJB-A & 25,808 & 500 & In-the-wild & Various & Various & 2D unconstrained face recognition\\ \hline
\end{tabular}
\end{adjustbox}
\end{center}
\caption{Comparison of datasets: UHDB31 \cite{Le_2017_17704} and IJB-A \cite{Klare_2015_17419}. Both are challenging due to pose variations, illumination, and resolution.}
\label{tab:cpr_dataset}
\end{table}

\subsection{Datasets}
UHDB31 \cite{Le_2017_17704} was created in a controlled lab environment, which allows face-related research on pose and illumination issues. 
In addition to 2D images, it also provides the corresponding 3D model of subjects.
An interesting fact of this dataset is that pose follows the uniform distribution on three dimensions: pitch, yaw, and roll.
For each subject, a total of 21 high-resolution 2D images from different views and 3D data are collected at the same time. 
Then, a 3D model is registered from the 3D data from different poses to generate a specific 3D face model.
In addition to three illuminations, the resolutions are downsampled to $128$, $256$, and $512$ from the original size.

IJB-A \cite{Klare_2015_17419} is another challenging dataset which consists of images in the wild.
This dataset was proposed by IARPA and is managed by NIST.
This dataset merges images and frames together and provides evaluations on the template level. 
A template contains one or several images/frames of a subject.
According to the IJB-A protocol, it splits galleries and probes into 10 folders.
In our experiment, we modify this protocol to use it for close-set face identification. The details will be introduced in Sec.~\ref{sec:exp_ijba}.
A summary of these two datasets is presented in Tab.~\ref{tab:cpr_dataset}.
Our system provides \texttt{dataset utility} to parse and load the data from these two datasets.

%\begin{figure*}[thb]
%\begin{center}
%%\fbox{\rule{0pt}{2in} \rule{0.9\linewidth}{0pt}}
%\includegraphics[width=\linewidth]{imgs/exp_rank1_uhdb31.pdf}
%\end{center}
%\caption{Comparison of Rank 1 of different systems on UHDB31.R128.I03.}
%\label{fig:rank1_uhdb31}
%\end{figure*}
\begin{table*}
\begin{center}
\begin{adjustbox}{max width=\linewidth}
\begin{tabular}{| c | c | c | c | c | c | c | c |}
\hline 
\diagbox{Pitch}{Yaw} & $-90^{\circ}$ & $-60^{\circ}$ &  $-30^{\circ}$ & $0^{\circ}$ & $+30^{\circ}$ & $+60^{\circ}$ & $+90^{\circ}$ \\ \hline
$+30^{\circ}$ &\begin{tabular}[c]{@{}c@{}}$14$/$11$/$58$/\\ $47$/$\mathbf{82}$\end{tabular} & 
\begin{tabular}[c]{@{}c@{}}$69$/$32$/$95$/\\ $90$/$\mathbf{99}$\end{tabular} & 
\begin{tabular}[c]{@{}c@{}}$94$/$90$/$\mathbf{100}$/\\$\mathbf{100}$/$\mathbf{100}$\end{tabular} & 
\begin{tabular}[c]{@{}c@{}}$99$/$\mathbf{100}$/$\mathbf{100}$/\\$\mathbf{100}$/$\mathbf{100}$\end{tabular} & 
\begin{tabular}[c]{@{}c@{}}$95$/$93$/$99$/\\$\mathbf{100}$/$99$\end{tabular}& 
\begin{tabular}[c]{@{}c@{}}$79$/$38$/$92$/\\$95$/$\mathbf{99}$\end{tabular}& 
\begin{tabular}[c]{@{}c@{}}$19$/$7$/$60$/\\$47$/$\mathbf{75}$\end{tabular}\\ \hline

$0^{\circ}$ & \begin{tabular}[c]{@{}c@{}}$22$/$9$/$84$/\\$81$/$\mathbf{96}$\end{tabular}& 
\begin{tabular}[c]{@{}c@{}}$88$/$52$/$99$/\\$\mathbf{100}$/$\mathbf{100}$\end{tabular}& 
\begin{tabular}[c]{@{}c@{}}$\mathbf{100}$/$99$/$\mathbf{100}$/\\$\mathbf{100}$/$\mathbf{100}$\end{tabular}& 
- & 
\begin{tabular}[c]{@{}c@{}}$\mathbf{100}$/$\mathbf{100}$/$\mathbf{100}$/\\$\mathbf{100}$/$\mathbf{100}$\end{tabular}& 
\begin{tabular}[c]{@{}c@{}}$94$/$73$/$99$/\\$100$/$\mathbf{100}$\end{tabular} & 
\begin{tabular}[c]{@{}c@{}}$27$/$10$/$91$/\\$84$/$\mathbf{96}$\end{tabular} \\ \hline

$-30^{\circ}$ & \begin{tabular}[c]{@{}c@{}}$8$/$0$/$44$/\\ $44$/$\mathbf{74}$\end{tabular}& 
\begin{tabular}[c]{@{}c@{}}$2$/$19$/$80$/\\ $90$/$\mathbf{97}$\end{tabular}& 
\begin{tabular}[c]{@{}c@{}}$91$/$90$/$99$/\\$99$/$\mathbf{100}$\end{tabular}& 
\begin{tabular}[c]{@{}c@{}}$96$/$99$/$99$/\\$\mathbf{100}$/$\mathbf{100}$\end{tabular}& 
\begin{tabular}[c]{@{}c@{}}$96$/$98$/$97$/\\$99$/$\mathbf{100}$\end{tabular} & 
\begin{tabular}[c]{@{}c@{}}$52$/$15$/$90$/\\$95$/$\mathbf{96}$\end{tabular}& 
\begin{tabular}[c]{@{}c@{}}$9$/$3$/$35$/\\$58$/$\mathbf{78}$\end{tabular}\\ \hline
\end{tabular}
\end{adjustbox}
\end{center}
\caption{Comparison of Rank-1 percentage of different systems on UHDB31.R128.I03. The methods are ordered as VGG-Face, COTS v1.9, FaceNet, \name -PRFS, and \name-DPRFS. The index of poses are ordered from the left to right and from the top to bottom (\eg, pose 3 is pitch $-30^{\circ}$ and yaw $-90^{\circ}$, pose 11 is  pitch $0^{\circ}$ and yaw $0^{\circ}$). The frontal face is gallery while the other poses are probes. In all cases, our system achieves the best performance compared with the state-of-the-art.}
\label{tab:rank1_uhdb31}
\end{table*}

\subsection{Baselines}
To perform a fair comparison with current state-of-the-art face recognition systems, we choose VGG-Face and COTS v1.9 as baselines.

The VGG-Face descriptor was developed by Parkhi \etal \ \cite{Parkhi_2015_16638}. 
The original release contains a Caffe model and a \mbox{MATLAB} example.
We re-used their model, implemented their embedding method on multi-scaled images, and fused the features in C++. 
In our implementation, we tried different combinations of descriptor and matching methods.
We found that embedding features with cosine similarity metrics works the best for the VGG-Face. 
In our experiment, we use VGG-Face to represent the embedding features with matching using a cosine similarity metric.
As in the baseline module, \name \ provides API to obtain the features.

The FaceNet algorithm was proposed by Schroff \etal \ \cite{Schroff_2015_16462}. We use a personal implemented FaceNet from GitHub~\footnote{\url{https://github.com/davidsandberg/facenet}} trained using WebFace \cite{Yi_2014_17446} and MS-Celeb-1M \cite{Guo_2016_17416}. 
They first use MTCNN \cite{Zhang_2016_17265} to align face and extract $128$ dimensions features. 
They provide pre-trained models that achieves $99.20\% \pm 0.30\%$ accuracy on the LFW dataset.
The accuracy is a little bit lower than the original paper, but still can be considered state-of-the-art.

COTS is a commercial software developed for scalable face recognition. It provides SDK and applications which can be used directly.
In our experiments, we used version 1.9 to compare with our system.
This version is considered to be a significant boost compared with previous versions.

In our experiment, we report the performance using both PRFS and DPRFS features. The summary of software configuration is reported in Tab.~\ref{tab:cpr_system_exp}. We compute the Rank-1 identity accuracy from successfully enrolled signatures.

\begin{table}[htb]
\begin{center}
\begin{tabular}{|l|c|c|c|}
\hline
System & Features & Dims & Metric \\ \hline \hline
VGG-Face & Embedding & 4096 & Cosine \\ \hline
COTS v1.9 & - & - & - \\ \hline
FaceNet  & - & 128 & Cosine \\ \hline
\name & PRFS & $64\times1024$ & Cosine \\ \hline
\name & DPRFS & $8\times1024$ & Cosine \\ \hline
\end{tabular}
\caption{Comparison of systems configuration used in our experiments.}
\label{tab:cpr_system_exp}
\end{center}
\end{table}

\subsection{UHDB31: Pose-Invariant Face Recognition}
\label{sec:exp_uhdb31}
%\subsubsection{Configuration}
In this experiment, we chose a configuration from the UHDB31 dataset named UHDB31.R0128.I03.
This is a subset in which all images are down-sampled to the size $153 \times 128$ in the neutral illumination.
This subset was chosen to demonstrate that our system, \name, is robust to different poses. 
Therefore, we use this configuration to exclude the other variations such as illumination, expressions, \etc, but only keep the pose variations.

We treated the frontal face images (pose 11) as gallery and images from the other 20 poses (poses $1-10$, $12-21$) as probes, independently.
Both the gallery and the probe contain $77$ images, each of which belongs to a subject.
The face identification experiment was performed using 20 pairs of sigsets.

%\subsubsection{Results}
Table~\ref{tab:rank1_uhdb31} depicts the comparison of Rank-1 accuracy among 20 poses (except pose 11, which is used for gallery), which indicates that  \name \ is robust to the different poses compared with other systems. 
We observed the VGG-Face and COTS v1.9 algorithms cannot generalize all pose distributions. 
FaceNet works better than VGG-Face and COTS v1.9 on the extreme poses. 
One possible answer is that this model is trained from the most available datasets using Ms-Celeb-1M and WebFace, 
which provide more extreme pose cases.
However in cases such as pose 3 ($-30^{\circ}, -90^{\circ}$) and pose 21 ($-30^{\circ}, -90^{\circ}$) in Tab.~\ref{tab:rank1_uhdb31}, the performance of 2D only face recognition pipelines still has significant room for improvement.
%With increasing poses, the performance from the other algorithms drop significantly.
On the other hand, with the help of the 3D model, our system keeps the consistent and symmetric performance among the different poses.
Even in the cases with yaw $-90^{\circ}$ or $+90^{\circ}$, our system can tolerate the pose variations, and achieves around $80\%$ Rank-1 identity accuracy with DPRFS features and around $50\%$ Rank-1 identity accuracy with PRFS features on average.
%We can conclude that our system is a pose-invariant face recognition system.
%In Fig. \ref{fig:roc_uhdb31}, we show a general improvement of \name compared with VGG-Face and COTS software.
%
%\begin{figure}[thb]
%\begin{center}
%\end{center}
%\caption{Depiction of COTS on UHDB31.}
%\label{fig:roc_uhdb31}
%\end{figure}
%
%\begin{figure}[thb]
%\begin{center}
%\end{center}
%\caption{Depiction of CMC on UHDB31.R128.I03.}
%\label{fig:cmc_uhdb31}
%\end{figure}

\subsection{IJB-A: In-the-Wild Face Recognition}
\label{sec:exp_ijba}

%\begin{figure*}[thb]
%\begin{center}
%%\fbox{\rule{0pt}{2in} \rule{0.9\linewidth}{0pt}}
%\includegraphics[width=\linewidth]{imgs/exp_rank1_ijba.pdf}
%\end{center}
%\caption{Comparison of Rank 1 of different systems on 10 splits of IJB-A. \name \ achieves the best performance on each split.}
%\label{fig:rank1_ijba}
%\end{figure*}

%\subsubsection{Configuration}
%\label{sec:ijba_config}
However, in a real-world case, a face recognition system does not suffer only from pose variations.
In this experiment, we want to explore whether our system is can also be used in an in-the-wild environment.
%Despite the original protocol from IJB-A dataset, we designed another face identification experiments based on the original 10 splits. 
We designed a different protocol for face identification experiments based on the original 10 splits.
Unlike the original template-level comparison, we conducted an image pairs comparison.
First, we removed some samples in the IJB-A splits to make 10 close-set comparison pairs.
Then, we cropped the face according to the annotations.
Image thumbnails with resolution below $50$ were up-sampled, while those with resolution larger than $1000$ were down-sampled. 
%We did $10$ face identification experiments on these splits.
Herein, we do not compare with FaceNet since there are overlapping samples between the training set and IJB-A dataset.

%\subsubsection{Results}
\begin{table*}[thb]
\begin{center}
\begin{adjustbox}{max width=\linewidth}
\begin{tabular}{|l|c|c|c|c|c|c|c|c|c|c|c|}
\hline
Method & Split-1 & Split-2 & Split-3 & Split-4 & Split-5 & Split-6 & Split-7 & Split-8 & Split-9 & Split-10 & Avg. \\ \hline \hline
VGG-Face & $76.18$ & $74.37$ & $24.33$& $47.67$ & $52.07$ & $47.11$ & $58.31$ & $54.31$ & $47.98$ & $49.06$ & $53.16$ \\ \hline
COTS v1.9 & $75.68$ & $76.57$ & $73.66$ & $76.73$ & $76.31$ & $77.21$ & $76.27$ & $74.50$ & $72.52$ & $77.88$ & $75.73$ \\ \hline
\name-PRFS & $47.61$ & $49.27$ & $47.71$ & $47.71$ & $48.97$ & $44.83$ & $52.98$ & $44.14$ & $43.40$ & $49.02$ & $47.56$ \\ \hline
\name-DPRFS & $\mathbf{78.20}$ & $\mathbf{76.97}$ & $\mathbf{77.31}$ & $\mathbf{79.00}$ & $\mathbf{78.01}$ & $\mathbf{79.00}$ & $\mathbf{81.15}$ & $\mathbf{78.40}$ & $\mathbf{74.97}$ & $\mathbf{78.57}$ & $\mathbf{78.16}$ \\ \hline
\end{tabular}
\end{adjustbox}
\end{center}
\caption{Comparison of Rank-1 percentage of different systems on 10 splits of IJB-A. \name \ achieves the best performance with DPRFS features on each split.}
\label{tab:rank1_ijba}
\end{table*}

Table~\ref{tab:rank1_ijba} depicts the rank-1 identification rate with different methods on IJB-A dataset.
Our system \name \ with DPRFS reports better performance compared with VGG-Face and COTS v1.9.
Also, our system results are consistent on 10 splits, which indicates that our system is robust.
Why do PRFS features in our system not perform well on the IJB-A dataset?
One possible answer is that PRFS features are trained on the FRGC dataset, which has notably fewer variations of pose, illumination, and resolution problems.
The current PRFS features cannot generalize on these images with large variances.
The corresponding solution is retraining the PRFS feature model on the in-the-wild dataset.
Third, COTS performs well on this challenging dataset, since it is designed for the real scenario.
%Finally, we come out a question that comparing the experiment in Sec. \ref{sec:exp_uhdb31}, why our system only outperforms slightly better than baselines?
By comparing the experiment in Sec.~\ref{sec:exp_uhdb31}, we are left with the question why does our system perform only slightly better than baselines?
We argue that in in-the-wild scenarios there are complicated combinations of pose variations, illumination, expression, and occlusions.
A robust face recognition system should take all cases into consideration.
In addition, COTS dropped hard samples and enrolled fewer signatures than ours, which would boost the performance to some extent.
%\begin{figure}[thb]
%\begin{center}
%\end{center}
%\caption{Depiction of COTS on IJB-A.}
%\label{fig:roc_ijba}
%\end{figure}
%
%\begin{figure}[thb]
%\begin{center}
%\end{center}
%\caption{Depiction of CMC on IJB-A.}
%\label{fig:cmc_ijba}
%\end{figure}

\begin{table}[htb]
\begin{center}
\begin{adjustbox}{max width=\linewidth}
\begin{tabular}{|l|c|}
\hline
Method & Rank-1 (\%) \\ \hline \hline
OpenBR \cite{Klontz_2013_14723}& $24.60\pm1.10$ \\
Wang \etal \ \cite{Wang_2017_17758} & $82.20\pm2.30$ \\
DCNN \cite{Chen_2016_17757} & $85.20\pm1.80$ \\
PAM \cite{Masi_2016_17266} & $77.10\pm1.60$\\
DR-GAN \cite{Tran_2017_17705} & $85.50\pm1.50$ \\
\hline
\name & $\mathbf{85.65\pm1.74}$ \\ \hline
\end{tabular}
\end{adjustbox}
\end{center}
\caption{Comparison of Rank-1 percentage of different systems on 10 splits of IJB-A. \name \ achieves the best performance with DPRFS features on each split.}
\label{tab:rank1_ijba_template}
\end{table}

\begin{table}[htb]
\begin{center}
\begin{adjustbox}{max width=\linewidth}
\begin{tabular}{|l|c|c|c|c|c|c|c|c|c|c|c|}
\hline
Method & Split-1 & Split-2 & Split-3 & Split-4 & Split-5 & Split-6 & Split-7 & Split-8 & Split-9 & Split-10 & Avg. \\ \hline \hline
VGG-Face & $74.44$ & $74.26$ & $70.68$& $73.96$ & $69.60$ & $72.64$ & $72.91$ & $70.03$ & $72.25$ & $71.78$ & $72.25$ \\ \hline
\name (DPRFS) & $\mathbf{87.22}$ & $\mathbf{86.82}$ & $\mathbf{83.68}$ & $\mathbf{86.05}$ & $\mathbf{83.52}$ & $\mathbf{88.22}$ & $\mathbf{85.14}$ & $\mathbf{83.59}$ & $\mathbf{84.86}$ & $\mathbf{87.38}$ & $\mathbf{85.65}$ \\ \hline
\end{tabular}
\end{adjustbox}
\end{center}
\caption{Detailed Rank-1 percentage of different systems on 10 splits of IJB-A.}
\label{tab:rank1_ijba_template_split}
\end{table}

We extended \name \ to enroll the several images for a subject to generate a template. The template is an average of signatures computed by generating a unified 3D model from several 2D images.
Here we use the results from \cite{Tran_2017_17705} to do the comparison.
Table~\ref{tab:rank1_ijba_template} lists the average Rank-1 identification accuracy for each method. 
\name \ achieved the best performance.
The detailed comparison of Rank-1 identification accuracy with VGG-Face is summarized in Tab.~\ref{tab:rank1_ijba_template_split} for 10 splits in the IJB-A dataset.

\subsection{Memory Usage and Running Time}

We conducted the analysis of \name \ in terms of both memory and time.
Caffe-related implementation runs on GPU (GTX TITAN X). 
COTS v1.9 makes full use of eight CPUs.
Table~\ref{tab:cpr_system_time} summarizes the system run-times for different systems.
Some modules in our implementation or external libraries run on CPU, such as face detection, pose estimation, text-lifting, and PRFS feature extraction.
Therefore, the time used by PRFS features takes $1.5$ s more than DPRFS features.
Due to loading several large models, DPRFS requires more memory.
The user can define the suitable feature extractors according to their needs.
Since we optimized Memory for DPRFS, it shares the memory block in the GPU.
The memory cost is reduced to the same level as PRFS.
We use DPRFS by default.

\begin{table}[htb]
\begin{center}
\begin{tabular}{|l|c|c|c|}
\hline
System & GPU & Memory (GB) & Time (s)\\ \hline \hline
VGG-Face & Full  & 1.2 & 0.9 \\ \hline
COTS v1.9 & No &  0.1 & 0.5 \\ \hline
\name\ (PRFS) & Partial & 2.4 &  2.5 \\ \hline
\name\ (DPRFS) & Partial & 2.4 & 1.0 \\ \hline
\end{tabular}
\caption{Comparison of system run-times. ``Partial" in GPU column denotes part of the code does not support GPU acceleration. Time means the average enrollment time for a single image.}

\label{tab:cpr_system_time}
\end{center}
\end{table}

%\section{Future Work}
%\label{sec:future}
%We are keeping updating the system.
%In next version, we are going to support following new features in \name:
%\begin{itemize}
%\item Template-level enrollment: Current \name \ does not support multiple image enrollment, which limits the horizontal comparison with other methods on IJB-A datasets;
%\item Discriminative features: We are going to train discriminative features on Ms-Celeb-1M, which is the largest face dataset available in the world;
%\item Modules updating: We are going to update the modules when the new algorithms are proposed.
%\end{itemize}

\section{Conclusion}
\label{sec:con}
In this paper, a well-designed 3D-aided 2D face recognition system (\name) that is robust to pose variations as large as $90^{\circ}$ using deep learning technology has been presented.
An overview of the architecture, interface, and each module in \name \ are introduced i detailed.
Extensive experiments are conducted on UHDB31 and IJB-A to demonstrate that \name \ is robust to the pose variations, and it outperforms existing 2D-only face recognition systems such as VGG face descriptor, FaceNet, and a commercial face recognition software by at least $9\%$ on UHDB31 dataset and $3\%$ on IJB-A dataset in average.
And the system achieves the state-of-the-art performance of $85\%$ in template matching on IJB-A dataset.

\section{Acknowledgment}
This material is based upon work supported by the U.S. Department of Homeland Security under Grant Award Number 2015-ST-061-BSH001. This grant is awarded to the Borders, Trade, and Immigration (BTI) Institute: A DHS Center of Excellence led by the University of Houston, and includes support for the project ``Image and Video Person Identification in an Operational Environment'' awarded to the University of Houston. The views and conclusions contained in this document are those of the authors and should not be interpreted as necessarily representing the official policies, either expressed or implied, of the U.S. Department of Homeland Security.

\section{References}
{\small
\bibliographystyle{ieee}
\bibliography{../../bi}

\begin{thebibliography}{10}
\expandafter\ifx\csname url\endcsname\relax
  \def\url#1{\texttt{#1}}\fi
\expandafter\ifx\csname urlprefix\endcsname\relax\def\urlprefix{URL }\fi
\expandafter\ifx\csname href\endcsname\relax
  \def\href#1#2{#2} \def\path#1{#1}\fi

\bibitem{Le_2017_17704}
H.~Le, I.~A. Kakadiaris, {UHDB}31: A dataset for better understanding face
  recognition across pose and illumination variation, in: Proc. IEEE
  International Conference on Computer Vision Workshops, Venice, Italy, 2017.

\bibitem{Klare_2015_17419}
B.~F. Klare, B.~Klein, E.~Taborsky, A.~Blanton, J.~Cheney, K.~Allen,
  P.~Grother, A.~Mah, M.~Burge, A.~K. Jain, Pushing the frontiers of
  unconstrained face detection and recognition: {IARPA janus benchmark A}, in:
  Proc. IEEE Conference on Computer Vision and Pattern Recognition, Boston,
  Massachusetts, 2015, pp. 1931--1939.

\bibitem{Taigman_2014_15190}
Y.~Taigman, M.~Yang, M.~Ranzato, L.~Wolf, Deep{F}ace: {Closing} the gap to
  human-level performance in face verification, in: Proc. IEEE Conference on
  Computer Vision and Pattern Recognition, Columbus, Ohio, 2014, pp. 1701 --
  1708.

\bibitem{Huang_2008_13354}
G.~B. Huang, M.~Mattar, T.~Berg, E.~Learned-Miller, {Labeled faces in the
  Wild}: {A} database for studying face recognition in unconstrained
  environments, in: Proc. Workshop on Faces in 'Real-Life' Images: Detection,
  Alignment, and Recognition, Marseille, France, 2008.

\bibitem{Schroff_2015_16462}
F.~Schroff, D.~Kalenichenko, J.~Philbin, Face{N}et: A unified embedding for
  face recognition and clustering, in: Proc. IEEE Conference on Computer Vision
  and Pattern Recognition, Boston, Massachusetts, 2015, pp. 815--823.

\bibitem{Yi_2014_17446}
D.~Yi, Z.~Lei, S.~Liao, S.~Z. Li, Learning face representation from scratch,
  ArXiv preprint arXiv:1411.7923 (2014) 1--9.

\bibitem{Guo_2016_17416}
Y.~Guo, L.~Zhang, Y.~Hu, X.~He, J.~Gao, {MS}-{C}eleb-1{M}: A dataset and
  benchmark for large-scale face recognition, in: Proc. $14^{th}$ European
  Conference on Computer Vision, Amsterdam, Netherlands, 2016, pp. 87--102.

\bibitem{Parkhi_2015_16638}
O.~M. Parkhi, A.~Vedaldi, A.~Zisserman, Deep face recognition, in: Proc.
  British Machine Vision Conference, Swansea, UK, 2015, pp. 1--12.

\bibitem{He_2016_17189}
K.~He, X.~Zhang, S.~Ren, J.~Sun, Identity mappings in deep residual networks,
  in: Proc. European Conference on Computer Vision, Amsterdam, the Netherlands,
  2016, pp. 1--15.

\bibitem{Wen_2016_17264}
Y.~Wen, K.~Zhang, Z.~Li, Y.~Qiao, A discriminative feature learning approach
  for deep face recognition, in: Proc. $14^{th}$ European Conference on
  Computer Vision, Amsterdam, Netherlands, 2016, pp. 499--515.

\bibitem{Zhang_2016_17759}
X.~Zhang, Z.~Fang, Y.~Wen, Z.~Li, Y.~Qiao, Range loss for deep face recognition
  with long-tail, ArXiv preprint arXiv:1611.08976 (2016) 1--9.

\bibitem{Liu_2017_17761}
W.~Liu, Y.~Wen, Z.~Yu, M.~Li, B.~Raj, L.~Song, Sphere{F}ace: deep hypersphere
  embedding for face recognition, in: Proc. IEEE Conference on Computer Vision
  and Pattern Recognition, Honolulu, Hawaii, 2017, pp. 212 -- 220.

\bibitem{Chen_2017_17762}
B.~Chen, W.~Deng, J.~Du, Noisy softmax: improving the generalization ability of
  dcnn via postponing the early softmax saturation, in: Proc. IEEE Conference
  on Computer Vision and Pattern Recognition, Honolulu, Hawaii, 2017, pp.
  5372--5381.

\bibitem{Kakadiaris_2017_13356}
I.~A. Kakadiaris, G.~Toderici, G.~Evangelopoulos, G.~Passalis, D.~Chu, X.~Zhao,
  S.~K. Shah, T.~Theoharis, {3D-2D} face recognition with pose-illumination
  normalization, Computer Visiona and Image Understanding 154 (2017) 137--151.

\bibitem{Hu_2016_17415}
G.~Hu, F.~Yan, C.~Chan, W.~Deng, W.~Christmas, J.~Kittler, N.~M. Robertson,
  Face recognition using a unified 3{D} morphable model, in: Proc. $14^{th}$
  European Conference on Computer Vision, Amsterdam, Netherlands, 2016.

\bibitem{Ding_2016_16547}
C.~Ding, D.~Tao, A comprehensive survey on pose-invariant face recognition, ACM
  Transactions on intelligent systems and technology 7~(3) (2016) 1--40.

\bibitem{noauthor__17451}
Open{CV}, \url{http://opencv.org}.

\bibitem{noauthor__17452}
Glog, \url{https://github.com/google/glog}.

\bibitem{noauthor__17453}
Gflags, \url{https://github.com/gflags/gflags}.

\bibitem{noauthor__17454}
Pugixml, \url{https://github.com/zeux/pugixml}.

\bibitem{noauthor__17455}
{JSON} for modern {C}++, \url{https://github.com/nlohmann/json}.

\bibitem{Jia_2014_16330}
Y.~Jia, E.~Shelhamer, J.~Donahue, S.~Karayev, J.~Long, R.~B. Girshick,
  S.~Guadarrama, T.~Darrell, Caffe: {C}onvolutional architecture for fast
  feature embedding, in: Proc. International Conference on Multimedia, Orlando,
  Florida, USA, 2014, pp. 675--678.

\bibitem{Xu_2017_17643}
X.~Xu, H.~Le, P.~Dou, Y.~Wu, I.~A. Kakadiaris, Evaluation of 3{D}-aided pose
  invariant 2{D} face recognition system, in: Proc. International Joint
  Conference on Biometrics, Denver, Colorado, 2017.

\bibitem{Zafeiriou_2015_17448}
S.~Zafeiriou, C.~Zhang, Z.~Zhang, A survey on face detection in the wild: past,
  present and future, Computer Vision and Image Understanding 138 (2015) 1--24.

\bibitem{Girshick_2014_17449}
R.~Girshick, J.~Donahue, T.~Darrell, J.~Malik, Rich feature hierarchies for
  accurate object detection and semantic segmentation, in: Proc. IEEE
  Conference on Computer Vision and Pattern Recognition, Columbus, OH, 2014,
  pp. 580--587.

\bibitem{Jiang_2017_17450}
H.~Jiang, E.~Learned-Miller, Face detection with the faster {R-CNN}, in: Proc.
  $12^{th}$ IEEE International Conference on Automatic Face \& Gesture
  Recognition, Washington, DC, 2017, pp. 650--657.

\bibitem{Li_2016_17296}
Y.~Li, B.~Sun, T.~Wu, Y.~Wang, Face detection with end-to-end integration of a
  {ConvNet} and a {3D} model, in: Proc. $14^{th}$ European Conference on
  Computer Vision, Amsterdam, Netherlands, 2016, pp. 420--436.

\bibitem{Hu_2017_17443}
P.~Hu, D.~Ramanan, Finding tiny faces, in: Proc. IEEE Conference on Computer
  Vision and Pattern Recognition, Honolulu, Hawaii, 2017, pp. 951--959.

\bibitem{He_2016_17161}
K.~He, X.~Zhang, S.~Ren, J.~Sun, Deep residual learning for image recognition,
  in: Proc. Computer Vision and Pattern Recognition, Las Vegas, NV, 2016, pp.
  770--778.

\bibitem{Liu_2016_17507}
W.~Liu, D.~Anguelov, D.~Erhan, C.~Szegedy, S.~Reed, C.~Fu, A.~C. Berg, {SSD}:
  single shot multibox detector, in: Proc. European Conference on Computer
  Vision, Amsterdam, Netherlands, 2016, pp. 21--37.

\bibitem{Redmon_2017_17763}
J.~Redmon, A.~Farhadi, {YOLO}9000: better, faster, stronger, in: Proc. IEEE
  Conference on Computer Vision and Pattern Recognition, Honolulu, Hawaii,
  2017, pp. 7263--7271.

\bibitem{Lin_2017_17764}
T.-Y. Lin, P.~Goyal, R.~Girshick, K.~He, P.~Dollar, Focal loss for dense object
  detection, ArXiv preprint arXiv:1708.02002 (2017) 1--10.

\bibitem{Najibi_2017_17765}
M.~Najibi, P.~Samangouei, R.~Chellappa, L.~S. Davis, {SSH}: single stage
  headless face detector, ArXiv preprint arXiv:1708.03979 (2017) 1--10.

\bibitem{Jin_2017_17152}
X.~Jin, X.~Tan, Face alignment in-the-wild: A survey, Computer Vision and Image
  Understanding (2017) 1--22.

\bibitem{Zhu_2015_16323}
S.~Zhu, C.~Li, C.~C. Loy, X.~Tang, Face alignment by {coarse-to-fine} shape
  searching, in: Proc. IEEE Conference on Computer Vision and Pattern
  Recognition, Boston, MA, 2015, pp. 4998--5006.

\bibitem{Xu_2016_16539}
X.~Xu, S.~Shah, I.~A. Kakadiaris, Face alignment via an ensemble of random
  ferns, in: Proc. IEEE International Conference on Identity, Security and
  Behavior Analysis, Sendai, Japan, 2016.

\bibitem{Xu_2017_17394}
X.~Xu, I.~A. Kakadiaris, Joint head pose estimation and face alignment
  framework using global and local {CNN} features, in: Proc. $12^{th}$ IEEE
  Conference on Automatic Face \& Gesture Recognition, Washington, DC, 2017,
  pp. 642--649.

\bibitem{Kumar_2017_17402}
A.~Kumar, A.~Alavi, R.~Chellappa, {KEPLER}: keypoint and pose estimation of
  unconstrained faces by learning efficient h-cnn regressors, in: Proc.
  $12^{th}$ IEEE Conference on Automatic Face \& Gesture Recognition,
  Washington, DC, 2017, pp. 258--265.

\bibitem{Wu_2017_17445}
Y.~Wu, S.~K. Shah, I.~A. Kakadiaris, Go{DP}: {Globally} optimized dual pathway
  system for facial landmark localization in-the-wild, Image and Vision
  Computing (2017) 1--16(Under review).

\bibitem{Huang_2017_17458}
R.~Huang, S.~Zhang, T.~Li, R.~He, Beyond face rotation: global and local
  perception gan for photorealistic and identity preserving frontal view
  synthesis, ArXiv preprint arXiv:1704.04086 (2017) 1--11.

\bibitem{Yin_2017_17457}
X.~Yin, X.~Yu, K.~Sohn, X.~Liu, M.~Chandraker, Towards large-pose face
  frontalization in the wild, ArXiv preprint arXiv:1704.06244 (2017) 1--12.

\bibitem{Tran_2017_17705}
L.~Tran, X.~Yin, X.~Liu, Disentangled representation learning {GAN} for
  pose-invariant face recognition, in: Proc. IEEE Conference on Computer Vision
  and Pattern Recognition, Honolulu, Hawaii, 2017, pp. 1415 -- 1424.

\bibitem{Masi_2016_17266}
I.~Masi, S.~Rawls, G.~Medioni, P.~Natarajan, Pose-aware face recognition in the
  wild, in: Proc. IEEE Conference on Computer Vision and Pattern Recognition,
  Las Vegas, NV, 2016, pp. 4838 -- 4846.

\bibitem{Masi_2016_17399}
I.~Masi, A.~Trần, T.~Hassner, J.~Leksut, G.~Medioni, Do we really need to
  collect millions of faces for effective face recognition?, in: Proc. European
  Conference on Computer Vision, Amsterdam, The Netherlands, 2016, pp.
  579--596.

\bibitem{Deng_2017_17766}
J.~Deng, Y.~Zhou, S.~Zafeiriou, Marginal loss for deep face recognition, in:
  Proc. IEEE Conference on Computer Vision and Pattern Recognition, Honolulu,
  Hawaii, 2017, pp. 60--68.

\bibitem{Klontz_2013_14723}
J.~Klontz, B.~Klare, S.~Klum, A.~Jain, M.~Burge, Open source biometric
  recognition, in: Proc. IEEE Conference on Biometrics: Theory, Applications
  and Systems, Washington DC, 2013.

\bibitem{Sun_2015_16639}
Y.~Sun, D.~Liang, X.~Wang, X.~Tang, Deep{ID}3: face recognition with very deep
  neural networks, arXiv preprint arXiv:1502.00873 (2015) 1--5.

\bibitem{Amos_2016_16583}
B.~Amos, B.~Ludwiczuk, S.~Mahadev, Open{F}ace: A general-purpose face
  recognition library with mobile applications, Tech. Rep. CMU-CS-16-118, CMU
  School of Computer Science, Pittsburgh, PA (2016).

\bibitem{Zhang_2016_17265}
K.~Zhang, Z.~Zhang, Z.~Li, Y.~Qiao, Joint face detection and alignment using
  multitask cascaded convolutional networks, IEEE Signal Processing Letters
  23~(10) (2016) 1499--1503.

\bibitem{King_2009_17112}
D.~E. King, Dlib-ml: A machine learning toolkit, Journal of Machine Learning
  Research 10 (2009) 1755--1758.

\bibitem{Farfade_2015_16581}
S.~S. Farfade, M.~Saberian, L.~Li, Multi-view face detection using deep
  convolutional neural networks, in: Proc. $5^{th}$ ACM on International
  Conference on Multimedia Retrieval, Shanghai, China, 2015, pp. 643--650.

\bibitem{Mathias_2014_17444}
M.~Mathias, R.~Benenson, M.~Pedersoli, L.~V. Gool, Face detection without bells
  and whistles, in: Proc. 13th European Conference on Computer Vision, Zurich,
  Switzerland, 2014, pp. 720--735.

\bibitem{Krizhevsky_2012_17120}
A.~Krizhevsky, I.~Sutskever, G.~E. Hinton, Imagenet classification with deep
  convolutional neural networks, in: Proc. Neural Information Processing
  Systems, Lake Tahoe, NV, 2012, pp. 1097--1105.

\bibitem{Dou_2017_17421}
P.~Dou, S.~K. Shah, I.~A. Kakadiaris, End-to-end {3D} face reconstruction with
  deep neural networks, in: Proc. IEEE Conference on Computer Vision and
  Pattern Recognition, Honolulu, Hawaii, 2017, pp. 1--10.

\bibitem{Dou_2015_16379}
P.~Dou, L.~Zhang, Y.~Wu, S.~K. Shah, I.~A. Kakadiaris, Pose-robust face
  signature for multi-view face recognition, in: Proc. International Conference
  on Biometrics: Theory, Applications and Systems, Arlington, VA, 2015, pp.
  1--8.

\bibitem{Lei_2014_16175}
Z.~Lei, M.~Pietikainen, S.~Li, Learning discriminant face descriptor, IEEE
  Transactions on Pattern Analysis and Machine Intelligence 36~(2) (2014)
  289--302.

\bibitem{Wang_2017_17758}
D.~Wang, C.~Otto, A.~K. Jain, Face search at scale, IEEE Transactions on
  Pattern Analysis and Machine Intelligence 39 (2017) 1122 -- 1136.

\bibitem{Chen_2016_17757}
J.-C. Chen, J.~Zheng, V.~M. Patel, R.~Chellappa, Unconstrained face
  verification using deep cnn features, in: Proc. Winter Conference on
  Applications of Computer Vision (WACV), Lake Placid, NY, USA, 2016.

\end{thebibliography}
}

\end{document}